%% file: IEEE_TOSC.tex
\documentclass[journal]{IEEEtran}
\usepackage{amsmath,amsfonts}
\usepackage{algorithm}
\usepackage{algpseudocode} % modern replacement
\usepackage{array}
\usepackage[normalem]{ulem}
\usepackage{subcaption}
\usepackage[caption=false,font=footnotesize]{subfig}
\usepackage{geometry}
\usepackage[table]{xcolor}
\usepackage{textcomp}
\usepackage{stfloats}
\usepackage{url}
\usepackage{verbatim}
\usepackage{graphicx}
\usepackage{tabularx,tabulary}
\usepackage{siunitx}
\definecolor{RED}{rgb}{1,0,0}
\usepackage{cite}
\usepackage{xcolor}
\newcommand{\commentblue}[1]{\textcolor{black}{#1}}
\begin{document}
%todo change the name a bit
\title{BIPPO: Budget-Aware Independent PPO
for Energy-Efficient Federated Learning Services}

\author{ \IEEEauthorblockN{ 
    Anna Lackinger,
    Andrea Morichetta,
    Pantelis A. Frangoudis, 
    Schahram Dustdar
    }    
    \vspace{2mm}

    \IEEEauthorblockA{Distributed Systems Group, TU Wien, Vienna, Austria
    \\\{a.lackinger, a.morichetta, p.frangoudis, dustdar\}@dsg.tuwien.ac.at} 
%\thanks{This paper was produced by the IEEE Publication Technology Group. They are in Piscataway, NJ.}% <-this % stops a space
%\thanks{Manuscript received April 19, 2021; revised August 16, 2021.}
}

\markboth{\textcolor{RED}{Accepted at IEEE Transactions on Services Computing. Latest version available at DOI: 10.1109/TSC.2026.3713497.}}%
{Shell \MakeLowercase{\textit{et al.}}: A Sample Article Using IEEEtran.cls for IEEE Journals}


\maketitle

\begin{abstract}
% \commentblue{abstract 100-200 words!!}
Federated Learning (FL) is a promising approach for training models in large-scale IoT systems, natively ensuring load distribution and privacy. However, FL is agnostic to infrastructure efficiency, which is a critical concern in resource-constrained environments.
Although Reinforcement Learning (RL)-based solutions improve client selection in FL, they still overlook infrastructure challenges, such as resource constraints and client churn. Moreover, they often lack practical applicability, as they rarely guarantee generalizability and energy efficiency.
To fill this gap, we propose BIPPO (Budget-aware Independent Proximal Policy Optimization), an energy-efficient multi-agent RL (MARL) solution that enhances FL performance. We empirically evaluate BIPPO on two image classification tasks in highly budget-constrained, non-IID settings, a challenging scenario for vanilla FL. BIPPO's redesigned $\epsilon$-greedy sampler increases accuracy, convergence speed, and the model's consistency compared to non-RL mechanisms, traditional PPO, and IPPO. In addition, BIPPO consumes only a negligible share of the budget, which remains constant as the number of clients increases, resulting in a training overhead up to three orders of magnitude lower than that of single-agent RL (SARL). 
Overall, BIPPO delivers a performant, stable, scalable, and sustainable solution for client selection in IoT-FL.
% Overall, BIPPO is a generalizable, transferable mechanism, whose pretrained policies help jump-start model performance by almost 10\% on a new task.
\end{abstract} 


\input{sections/Introduction}
\input{sections/RW}
\input{sections/Sysmodel}

\input{sections/BIPPO}
\input{sections/Experiments}
\input{sections/Discussion}

\bibliographystyle{IEEEtran} 
\bibliography{bibliography}

\input{sections/bibliographies}

\vfill

\end{document}

%% file: sections/Introduction.tex
\section{Introduction}
\label{sec:intro}
Internet of Things (IoT) devices embed increasingly sophisticated sensing, computing, and connectivity capabilities that pervade our daily lives, leading to a surge in data generated at the edge.
% The proliferation of recent technical advances has led to an increasing number of devices embedding computational capabilities.
% The ubiquity of these Internet of Things (IoT) devices in our daily lives has led to a surge in data created at the edge.
To identify hidden patterns in the vast amount of data, big data analytics, machine learning (ML), and deep learning (DL) algorithms are essential~\cite{deng2021auction}.
The application of AI for data processing in IoT is also known as \textit{Artificial Intelligence of Things (AIoT)}, which is a new computing paradigm~\cite{awaisi2024survey}.
A leading solution for AIoT and for processing the vast amount of edge data is \textit{Federated Learning (FL)}~\cite{mcmahan2017communication}, due to its ability to distribute ML model training load while guaranteeing privacy and data locality.
%Fl problems and challenges
Despite its advantages, training shared models with FL comes with several challenges~\cite{zhang2022multi}. 
Without care, non-independent and identically distributed (non-IID) training data on client devices, characteristic of many IoT scenarios we target, can negatively affect FL convergence~\cite{wang2020optimizing,gao2024federated}.
Furthermore, as resources are finite, %it is crucial to properly orchestrate the ML services that use them~\cite{HEIDARI2022104089}. In the context of FL in particular, 
a major orchestration dimension is client selection~\cite{fu2023client}. 
Identifying a client selection strategy that autonomously balances the multidimensional problem of non-IID data and heterogeneous device resources is not trivial.
Several practical heuristic approaches have been proposed in recent years, but they cannot easily adapt to dynamic changes, and a fixed rule may lead to a biased selection and the overrepresentation of a few clients~\cite{fu2023client}.

%why we need DRL
In such complex scenarios, \textit{Reinforcement Learning (RL)}, particularly its deep variant (DRL), offers a novel perspective by enabling autonomous learning of effective strategies~\cite{azhati2024optimizing}.
However, no existing RL-based selector simultaneously guarantees a hard energy budget constraint, supports client churn, and keeps its own training cost independent of the number of clients (Table~\ref{table:FL_RL_summar}).
%budget + RL training energy
%clients joining/leaving
%
The main challenges we address in our work can be split into \textit{AI challenges}, related to model performance and generalizability, and \textit{IoT challenges}, related to infrastructure and resource management.

The AI challenges concern both our FL use case and our RL method.
The FL challenge is client selection in environments with \textit{non-IID data} and \textit{heterogeneous devices}.
Although various methods have been proposed to improve performance with non-IID data, no single best solution has emerged.
This FL challenge can be addressed using an RL method that autonomously learns an effective client selection policy.
However, many proposed RL-based solutions are not \textit{generalizable} to new environments as they are often trained to fit only one specific environment \cite{Mao2024joint,yan2023joint,yu2022proximal,zheng2024fedaeb}, making RL unapplicable in real-world scenarios. 
We aim at overcoming this limit by making sure that the RL strategy works even when data distribution or device capabilities information is not available. %, making it impossible to reproduce the exact training conditions.
%Therefore, making our RL-based selection generalizable to new scenarios is an important challenge we address.

Making our method \textit{sustainable} requires efficient resource management, a key challenge in IoT.
To support practical applications, a proposed method should operate efficiently in a budget-constrained environment, where available resources are limited.
Such budget-constrained orchestration scenarios can arise when coordinating multiple FL and/or other ML workflows in an as-a-service manner~\cite{gao2024federated,vcilic2024reactive} over shared edge-compute infrastructure.
In most existing RL work, adding such a budget constraint is not straightforward, as it heavily influences exploration. 
A low budget will lead clients to explore non-participation more, which biases their action preferences.
An important resource management aspect, often overlooked yet highly relevant in energy optimization problems, is the energy consumed by the RL method itself and how to minimize it~\cite{zheng2024fedaeb,Mao2024joint,yu2022proximal,yan2023joint}.
The importance of energy-efficient solutions is further emphasized by the growing number of industrial companies investing in environmental protection to meet increasing sustainability expectations~\cite{koliopoulos2022urban}.
%The sustainable use of resources is indeed crucial and has to be taken into account when finding solutions for efficient data processing.
The \textit{scalability} concern is crucial and, in our context, has two dimensions. The first one is related to maintaining consistent FL performance as the client population, data, and demand grow. The second one is the energy overhead of the client selection method itself, i.e., the sustainability problem above, which must not grow as the number of clients increases. %Therefore, we need to find methods that can easily adapt to expanding environments.
\textit{Stability} under client churn, i.e., when clients join or leave during FL training, is another important IoT challenge that has so far not been considered in most existing RL-based FL client selection schemes.
Due to structural changes, network availability, and device updates, FL clients may join or leave over time, and it is essential that the client selection method adapts to these dynamic environments.

To address these limitations, we propose \textit{Budget-aware Independent Proximal Policy Optimization (BIPPO)}, a novel approach for efficient client selection in FL.
BIPPO is based on IPPO, but we add a redesigned $\epsilon$-greedy sampler that enforces client selection under a hard budget constraint, while also improving overall performance.
%non-iid and heterogeneous devices + resource management
BIPPO is built to work in resource-constrained environments by intelligently selecting a subset of clients to avoid overrunning the limited energy budget, while considering multiple client parameters such as energy usage, data size, and local model performance.
Due to our energy-optimized training, BIPPO's training cost remains independent of the number of clients and is up to three orders of magnitude lower than that of single-agent RL (SARL) with 1,000 clients.
In summary, we make the following contributions:
\begin{itemize}
    \item \textit{New energy-aware client selection strategy~(\S\,\ref{sec:bippo})}: We introduce a novel RL method that coordinates a shared energy budget between various heterogeneous IoT devices in the context of FL~(\S\,\ref{sec:problem}).
    Contrary to other client selection methods~(\S\,\ref{sec:rwork}), \emph{the number of selected clients is not assumed fixed} but depends on the given budget and the energy consumption of the participating clients' devices. 
    \item \textit{Performance considering AI challenges (\S\,\ref{sec:evaluation})}: 
    Our extensive evaluation demonstrates that BIPPO, compared to traditional PPO, IPPO, and heuristic baselines, improves model accuracy and stabilizes performance in non-IID environments, ranking best or second best in mean accuracy and mean rounds needed to reach the target across all settings (Table~\ref{tab:summary_exp}). 
    It functions well in resource-constrained, low-budget environments, where identifying the right clients has a high impact on performance, and it \textit{generalizes} well to new datasets, where its pretrained policies provide a jump-start of up to $\sim$10\% accuracy.

    \item \textit{Performance considering IoT challenges (\S\,\ref{sec:evaluation})}: 
    BIPPO \textit{scales} along both the performance and the energy dimension in more complex environments. BIPPO maintains model accuracy in larger settings, while the training energy overhead remains constant in the number of clients, 14$\times$ lower than single-agent RL at 20 clients, up to three orders of magnitude at 1,000 clients, further highlighting the \textit{sustainability} of our design. BIPPO also shows consistent performance when clients join or leave, showing its \textit{stability} in dynamic environments.
    The implementation of our method and of our evaluation framework is available on \commentblue{GitHub
    \footnote{\url{https://github.com/Lacki28/BIPPO/}}}.

\end{itemize}

%% file: sections/RW.tex
\section{Related Work}
\label{sec:rwork}

\begin{table*}[ht]
\centering
\caption{FL non-IID client selection solutions in the literature that use RL.}
\label{table:FL_RL_summar}
\begin{tabularx}{1\textwidth}{|X|l l| c c |c  c c|}
\hline
\textbf{Ref.} & \textbf{Year} &  \textbf{RL Method} & \textbf{Device Heterogeneity/} &\textbf{Generalizability} & \textbf{Resource Limits}  & \textbf{Scalability} & \textbf{Stability} \\
& & & \textbf{non-IID Data}& \textbf{of RL method}&(\textbf{Budget}) &  & (\textbf{Client-churn})\\
\hline

\textbf{\cite{deng2021auction}}& 2022 & REINFORCE  & $\checkmark$  & $\sim$  & $\checkmark$  & $\checkmark$ &$\sim$\\
\textbf{\cite{chen2024towards}}& 2024 & DDPG  & $\checkmark$ &  $\sim$ & $-$  & $\sim$&$-$\\ 
\textbf{\cite{zhang2022multi}}& 2022 & VDN  & $\checkmark$ &$\checkmark$  & $-$  & $\checkmark$&$\sim$\\ 
\textbf{\cite{zhao2025fedppo}}& 2025 & PPO &$\checkmark$  & $\checkmark$ &  $-$ &$\sim$ &$-$\\ 
\textbf{\cite{Mao2024joint}}& 2024 &  REINFORCE & $\checkmark$ & $-$   &$-$ & $\sim$&$-$\\ 
\textbf{\cite{yan2023joint}}& 2023 & IPPO  & $\checkmark$ &$-$  & $-$  & $\sim$ & $\sim$\\

\textbf{\cite{wang2020optimizing,LiClientSelection2024}}& 2020,2024 & DDQN & $-$ &  $\sim$& $-$  & $-$ &$-$\\

\textbf{\cite{yu2022proximal}}& 2022 & PPO & $\checkmark$  & $-$   &$-$ &$-$ &$-$\\ 

\textbf{\cite{zheng2024fedaeb}}& 2023 & SAC & $\checkmark$ & $-$ & $-$  & $-$ &$-$\\ 

\hline
\textbf{BIPPO}& 2025 & IPPO & $\checkmark$ & 
$\checkmark$ & $\checkmark$  & $\checkmark$ 
& $\checkmark$ \\ 
\hline
\end{tabularx}

\end{table*}

\noindent\textbf{FL Solutions for Non-IID Data:}
In non-IID settings, data distribution and characteristics differ between clients, which can significantly affect FL performance, as the imbalanced class distribution leads to uneven local model updates. This, in turn, increases communication overhead~\cite{lu2024federated}, since the model takes longer to reach a satisfactory accuracy.
In recent years, various solutions have been proposed to mitigate the negative effects of non-IID data, which can be grouped into data-based, algorithmic, and system-level approaches~\cite{zhu2021federatedlearningnoniiddata}.

\textit{Data-based} solutions can be split into data augmentation and data sharing techniques.
Both of these approaches enhance learning performance, but most of them require data sharing, increasing the risk of data privacy leakage~\cite{zhu2021federatedlearningnoniiddata}.

A \textit{system-level} approach is to cluster the clients based on their data distribution.
In the work of Chen et al.~\cite{chen2024heterogeneity}, the FL server estimates the statistical heterogeneity of the data of each client by using the client’s update of the network’s output layer.
Based on this information, the clients are then clustered.
This can also be extended by not only taking the data distribution into account, but also classifying the clients based on additional factors, such as sample size~\cite{rai2022client}.
However, system-level approaches may need reconfiguration when the number of clients changes, as shown in~\cite{vcilic2024reactive}.
Reconfiguration may be expensive and delay FL training.

The selection of FL clients, which falls under both \textit{system-level} and \textit{algorithmic} approaches, is an efficient way to deal with non-IID data, as it can speed up convergence and increase accuracy~\cite{cho2022towards,fu2023client}. 
%To solve the challenge of choosing the right subset, various client selection algorithms have been developed that show promising performance improvements~\cite{fu2023client}.
However, determining the optimal client selection method for a specific FL process can be challenging due to the wide range of available strategies and influencing factors~\cite{mayhoub2024review}.
One algorithmic approach is to identify non-IID clients by looking at their weight divergence~\cite{zhang2021client}. 
Clients with a lower degree of non-IID data will more frequently be chosen for the training, which can significantly increase accuracy.
Another well-known method is the power-of-choice method, where clients with the highest loss are chosen to participate in the FL training~\cite{cho2020client}.
%However, these approaches do not focus on multiple dimensions, such as dataset size and device heterogeneity.

Several solutions have improved the overall performance in non-IID environments, yet these methods address a single dimension, e.g., data distribution, leaving out other important factors, such as device heterogeneity and dataset size.
Additionally, an algorithmic approach with a fixed selection strategy may not perform well in every environment, and it has no chance of learning or improving its strategy over time.
Due to these limitations, we focus on RL methods, since they can autonomously learn and improve their strategy in various environments.
With the appropriate reward configuration, RL can extend its utility by learning to trade off multiple objectives at once.
%which provides feedback for the RL agent to improve its policy, RL is also able to learn a strategy that considers multiple dimensions and autonomously learns a solution that finds a trade-off between multiple defined goals.

\noindent\textbf{RL for FL Client Selection} 
%To achieve an intelligent client selection that can learn and improve its strategy, several works have been presented that use RL to optimize FL.
Several works use RL to optimize FL client selection.
An overview of existing approaches is given in Table \ref{table:FL_RL_summar}, where we compare our method to existing RL methods that aim to learn an optimal client selection in heterogeneous environments.
A check mark in the table indicates that a feature is addressed in the paper, a minus sign that it is absent, and a tilde that it is either unclear or potentially applicable but not evaluated.
When considering the RL methods that have been used, one can observe that numerous approaches have been applied. However, no specific RL approach has yet become the state of the art in FL client selection.
Most methods consider not only non-IID data but also device heterogeneity.
Only about half of the RL methods have been evaluated in a new environment different from the one used for training. 
While existing work typically selects a fixed number of clients, only~\cite{deng2021auction} considers a learning budget for client selection.
In fact, while a few contributions tackle scalability and test their approach in different environments, none consider client churn.

To the best of our knowledge, our article presents the first RL-based client selection work that shows the effect of clients joining and leaving during the FL training process on performance, and that optimizes the RL training energy consumption.
In addition, our method is the first that considers a shared budget while not relying heavily on accurate pretraining, as it learns fast and is capable of adjusting to a changing number of clients.

%% file: sections/Sysmodel.tex
\section{System Model} 
\label{sec:problem}
In this section, we provide an overview of the system model, its components, and the information flow.
\subsection{System Model and Federated Learning Process}
\begin{figure}[htbp]
\centering
\includegraphics[width=0.5\textwidth]{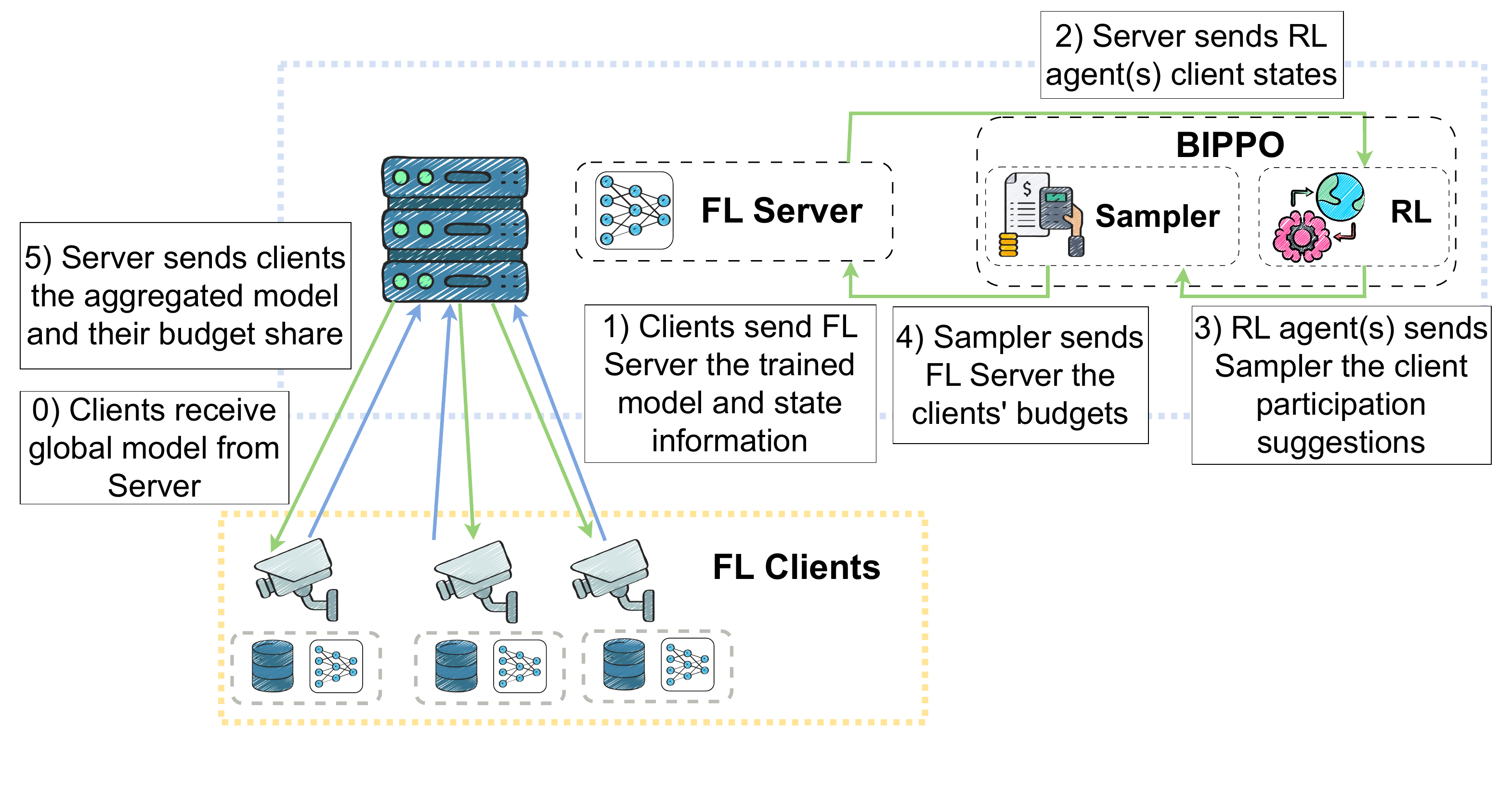}
\caption{Overview of our system design.}
\label{fig:ra}
\end{figure}

Figure \ref{fig:ra} illustrates the coordinated FL service with its elements and how they communicate with each other. 

At the core of our system is the FL process, which is a distributed model training technique. In each training round $t$, a shared global model is adjusted by aggregating local updates coming from the participating clients.
In detail, the FL process starts with the server sending the global model to the clients (Step 0).
After every participating client receives the most recent global model parameters \( \mathbf{w}^{t} \), each client \( n \) updates its local model parameters \( \mathbf{w}_n^{t} \) with the global model \( \mathbf{w}^t \) using gradient descent to minimize its local loss function \( L_n(\mathbf{w}_n^t) \), which is defined as the average of the individual batch losses \( l_j(\mathbf{w}_n^t) \)~\cite{deng2021auction}:
\begin{equation}
L_n(\mathbf{w}_n^t) = \frac{1}{|D_n|} \sum_{j \in D_n} l_j(\mathbf{w}_n^t) 
\end{equation}

At the beginning of a new FL workflow (round $t=0$), the clients train for only one epoch, and the updated local model parameters \( \mathbf{w}_n^{t+1} \) are sent to the server together with their state information. Simultaneously, each client also reports its estimated per-round energy cost, updated only if its device conditions change (Step 1). 
As it receives the updates, the server aggregates them using the Federated Averaging (FedAvg) algorithm:
\begin{equation}
\mathbf{w}^{t+1} = \frac{\sum_{n \in P^t} |D_n| \mathbf{w}_n^{t+1}}{\sum_{n \in P^t} |D_n|}
\end{equation}
where \( |D_n| \) is the number of data samples used by client \( n \) for local training and \( P^t \) is the set of clients participating in round \( t \).

In parallel with the aggregation, through BIPPO, we segment the selection phase into smaller steps (Steps 2--4) that help the central server choose the subset of clients that will train in the following round.
First, the server sends the RL agent(s) the state information of the clients (Step 2).
The RL agent(s) use their own model to evaluate possible actions in the given state and send them to the sampler (Step 3). In our model, we call these values, derived from the RL actor network(s), \emph{participation suggestions}.
More details on the sampler can be found in Section \ref{sec:sampler}.
The sampler decides on the final action, based on the participation suggestions and budget, and sends its decision to the server (Step 4).
The server then sends each client its budget share for the following round, together with the updated global model \( \mathbf{w}^{t+1} \) (Step 5).
The clients whose budget share is equal to or higher than their training cost can train locally and send their results back to the server (Step 1), which completes the round.

\subsection{Energy Consumption Model}
\label{sec:energy-model}
We model the energy cost of one FL round as the sum of the communication and computation energy of the participating clients.
Our model is consistent with previous work~\cite{chen2024towards,zhang2022energyefficient,10552399, liu2020client,Tran19}: it only takes into account the variable energy consumption, i.e., the energy consumed by the participating clients, and neglects other aspects of the FL process.

The first of the two terms is the \textbf{communication energy} \cite{liu2020client}, i.e., the energy a client requires to send its model update in one FL round.
% Since the downlink bandwidth is much larger than that of the uplink, the model downloading time is negligible as compared with the uploading time \cite{Mao2024joint}.
To make our work comparable to related work, for the communication energy, we use the formula proposed by \cite{Mao2024joint}.
% For the energy consumption we focus on the client side and the energy cost of transmitting the model and we ignore reception cost in line with \cite{Mao2024joint, federatedle}
Specifically, we can express the communication energy as: 
\begin{equation}
    E_n^{t,\text{comm}} =P_n  \frac{s_n}{B \log_2\left(1 + \frac{h_n P_n}{\sigma} \right)}
\end{equation}

\noindent where $B$ is the bandwidth, $P_n$ the transmitter power, $\sigma$ the background noise power, $h_n$ the channel gain, and $s_n$ the model size.

The second term in the energy consumption formula is the \textbf{computation energy}. We use the following expression to model it, which is typical in FL literature~\cite{liu2020client,10552399,Tran19}:
\begin{equation}
    E_n^{t,\text{comp}} = \frac{\alpha_n}{2} \, \eta_n \, |\tilde{D}_n| \, f_n^2
\end{equation}

\noindent where $|\tilde{D}_n|$ is the data size (in bits) that client $n$ trains on in round $t$, $f_n$ the CPU-cycle frequency, $\eta_n$ the number of CPU cycles to process one bit, and $\frac{\alpha_n}{2}$ the effective capacitance of client $n$’s computing chipset.

Given the communication and computation energy terms, we can express the energy consumed by one participating client as:
\begin{equation}
    E_n^{t,\text{total}} = E_n^{t,\text{comm}} + \text{epochs} \times E_n^{t,\text{comp}}
\end{equation}

The total energy consumed by all participating clients in one round, which should stay within the energy budget, can therefore be formulated as:
\begin{equation}
    E^{t,\text{total}} = \sum_{n \in P^t \subseteq N} \left( E_{n}^{t,\text{comm}} + \text{epochs} \times E_{n}^{t,\text{comp}} \right)
\end{equation}
where $P^t$ is the set of participating clients in round $t$, i.e., a subset of all clients $N$.

In our proposed framework, the FL clients use this function to estimate their own energy consumption and send their energy estimate to the server at the beginning of the FL process.
In a real-world deployment, these estimates could be replaced by energy measurements where available, as discussed in Section \ref{sec:practical-considerations}.

%% file: sections/BIPPO.tex
\section{BIPPO: Budget-aware Independent PPO}
\label{sec:bippo}
The main premise of BIPPO is that RL-based methods can autonomously learn to find a good strategy for client selection and adapt over time.
In RL, an agent learns a policy for selecting actions that lead to favorable subsequent states through interaction with the environment.
RL is a popular approach for FL client selection~\cite{deng2021auction,LiClientSelection2024,wang2020optimizing,zhao2025fedppo,Mao2024joint}, since it can learn complex patterns without having to define rules a priori. %, and it can adjust its strategy over time.
A popular RL method for FL client selection is Double Deep Q-Networks (DDQN)~\cite{van2016deep}, as shown by \cite{LiClientSelection2024,wang2020optimizing}. However, the effectiveness of DDQN can be inconsistent due to its sensitivity to hyperparameter choices, as demonstrated by Sadiki et al.~\cite{sadiki2023deep}, who showed that batch size alone can make a substantial performance difference.
As an alternative, we investigate Proximal Policy Optimization (PPO)~\cite{schulman2017proximal}.
Compared to DDQN, PPO is an on-policy method that does not have a replay buffer.
%Therefore, the data collected during its training does not need to be stored for long; once it has been used to update the policy, it is discarded.
This also makes PPO well-suited to our scenario, where we want to train the RL method at runtime without depending on large amounts of stored experience.
In addition, PPO is very effective in cooperative multi-agent settings~\cite{yu2022surprising}.
Yu et al.~\cite{yu2024multi} also recently showed that PPO and Multi-Agent PPO (MAPPO), in particular, outperform DDQN.
%PPO's effectiveness and robustness represent key desirable properties for our problem.
Compared to MAPPO, \emph{Independent PPO (IPPO)}~\cite{dewitt2020independentlearningneedstarcraft} is a logically decentralized design where each agent has its own actor and critic networks, and the agents do not exchange local information.
MAPPO uses a critic that takes into account the global state of all agents.
%whereas in IPPO, each agent has its own critic network.
When a new client joins, this influences the dimension of the global state space, and MAPPO would therefore need to train a new critic network. IPPO does not have this problem, since each agent has its own actor/critic networks, and their dimensions do not depend on the global state space.
% \subsection{Proximal Policy Optimization}
\commentblue{
PPO uses two neural networks, a policy network \( \pi_{\theta}(a|s) \) that outputs a probability distribution over the actions and a value network \( V_{\phi}(s) \) to estimate returns.
A key feature of PPO is the clipped surrogate objective function, which prevents large policy updates and stabilizes performance~\cite{schulman2017proximal,yan2023joint}. }

After a given batch size, based on data collected in the last rounds, the critic network is used to calculate the generalized advantage estimation (GAE)~\cite{schulman2017proximal,yan2023joint}.
For each agent $i$, the GAE can be calculated using the following formula:
\begin{equation}
A_t^i = \sum_{l=0}^{T} (\gamma \mu)^l \delta_{t+l}^i
\end{equation}
where $\delta_t^i = r_t^i + \gamma V_\phi(s_{t+1}^i) - V_\phi(s_t^i)$ is the Temporal Difference error at time step $t$. $V_\phi(s_t^i)$ and $V_\phi(s_{t+1}^i)$ denote the estimated state-value function at time steps $t$ and $t+1$, using the critic network $V_{\phi}$. $\gamma$ is the discount factor that controls the weighting of future rewards, and $\mu$ denotes the parameter used to govern the bias-variance trade-off in the estimation of the advantage function.

To stabilize training and prevent large, destabilizing updates to the policy, PPO optimizes a clipped surrogate objective function to calculate the policy loss for each agent $i$:
\begin{equation}
\begin{gathered}
L^{\text{CLIP}}(\theta_i) = \\
\mathbb{E}^i_t \left[ \min \left( r^i_t(\theta_i) \hat{A}^i_t, \text{clip}(r^i_t(\theta_i), 1 - \epsilon_{\text{clip}}, 1 + \epsilon_{\text{clip}}) \hat{A}^i_t \right) \right] 
\end{gathered}
\end{equation}

where \( r^i_t(\theta_i) = \frac{\pi_{\theta_i}(a^i_t|s^i_t)}{\pi_{\theta_i^{\text{old}}}(a^i_t|s^i_t)} \) is the probability ratio between the new and old policies, \( \hat{A}^i_t \) is the advantage estimate of agent $i$, 
% $\mathcal{C}$ is the clipping function $\text{clip}(r^i_t(\theta_i), 1 - \epsilon_{\text{clip}}, 1 + \epsilon_{\text{clip}})$
and \( \epsilon_{\text{clip}} \) is a hyperparameter controlling the trust region.

Additionally, the critic network parameters of each agent $i$ \( {\phi_i} \) are updated by minimizing the squared error between the predicted value and the observed return: 
\begin{equation}
  L^{\text{VF}}(\phi) = \mathbb{E}_t \left[ \left( V_{\phi_i}(s^i_t) - R^i_t \right)^2 \right]  
\end{equation}
where the return is defined as $R^i_t=A^i_t+V_{\phi^{old}_i}(s^i_t)$

% \subsection{IPPO}
% \emph{Independent PPO (IPPO)} approach~\cite{dewitt2020independentlearningneedstarcraft} is a logically decentralized design where each agent has its own value and critic network, and they do not exchange local information. 
% In comparison, MAPPO (Multi-Agent PPO) uses a critic that takes into account the global state, whereas in IPPO, each agent has its own critic network.
% When a new client joins, this influences the dimension of the global state space, and MAPPO would therefore need to retrain a new critic network. IPPO does not have this problem, since each agent has its own actor/critic networks, and their dimensions do not depend on the global state space.
% Furthermore, it should also be noted that with the IPPO design, the RL agent instances can be deployed either on the server or the client side, depending on the system requirements.
\subsection{Multi-Agent Reinforcement Learning}
To make our RL approach adapt to a changing number of clients, we formulate it as a cooperative \textit{Multi-Agent RL (MARL)} solution.
In cooperative MARL, a set of $I$ agents is trained to produce optimal actions that maximize \emph{team} reward~\cite{zhang2022multi}.  
Each client $n$ is paired with one agent; at each time step $t$, the agent observes the local state $s^t_{n}$ of its client and chooses an action $a^t_{n}$ based on that state.
After all agents act, the team receives a joint reward $r^t$, and the environment transitions to the next state $s^{t+1}_{n}$. The overall objective is to maximize the total expected reward.  

\noindent\textbf{State}: 
The state $s^t_{n}$ represents the state of client \(n\) at time $t$.
It is defined as $s^t_{n} = \{\overline{acc}^t,|D_n|, acc^t_{n}, e_{n},p_n^{t}\}$.
$s^t_{n}$ contains context information, which in this case is the average accuracy $\overline{acc}^t$ achieved after communication round $t$.
In addition, the state also consists of the dataset size of the client $|D_n|$, the accuracy on its local test dataset $acc^t_{n}$, the energy consumption for one training round $e_{n}$, and whether client $n$ participated in the last round $p_n^{t}$.

\noindent\textbf{Action}: 
In our FL setting, we consider $N$ clients. At each time step $t$, the RL agent(s) suggest for each client $n$ whether to participate in the FL training, denoted by a probability value $a^{t}_n$.
The participation suggestion is forwarded to the sampler, which will decide on the final action $\bar{a}^{t}_{n}$.
The final action is structured as a list including the budget share for each client in round $t$.
Details on the sampling strategy can be found in Section~\ref{sec:sampler}.

\noindent\textbf{Reward}: 
The reward function is used to evaluate the effectiveness of the action $a^t$ taken in state $s^t$. We define it as: 
$r(s^t, a^t)=
\varphi\lambda^{|\overline{acc}^t-\overline{acc}^{t-1}|}$, 
where $\lambda$ is a constant that makes the reward grow with the change in global test accuracy. 
\commentblue{Since the accuracy improvement of the global model tends to slow down as it approaches convergence, the updates become smaller.
To ensure that the DRL agent can still capture these small improvements, we set $\lambda$ to 64 in our experiments, which is consistent with related approaches \cite{wang2020optimizing,Mao2024joint,11017685}.}
Furthermore, $\varphi = \operatorname{sgn}(\overline{acc}^t-\overline{acc}^{t-1})$, i.e., attributes a positive sign to the reward if the model performance improved, negative otherwise.
% \sout{Inspired by related work \cite{wang2020optimizing}, we set $\lambda$ to 64.}
%When the last global accuracy is smaller than the current global accuracy, $\varphi$ is 1; if the accuracy has not changed, it is 0; otherwise, it is -1 to decrease the participation probability of the clients that took part in that round.
%Details on why the reward was defined this way and how this definition helps with the energy-aware RL training can be found in Section \ref{sec:training}.
The rationale is that, without $\varphi$ and the norm on the exponent, an exponential reward $\lambda^{(\overline{acc}^t-\overline{acc}^{t-1})}$ would remain positive even when accuracy drops (e.g., $64^{-0.5} = 0.125$), still providing a positive reinforcement to the selection of clients, even when they contribute to deteriorating the performance. We therefore remove the sign from the exponent and reapply it through $\varphi$, so that an accuracy drop leads to a penalty of the same magnitude, but opposite sign, as the reward for an equivalent gain.

The reward does not include the energy cost of the clients, since our goal is not to minimize energy consumption of the FL clients, but to optimize performance in energy-constrained environments.
% It may be the case that an optimal selection includes only one client, which takes up all the budget, but it has a balanced dataset that helps improve performance.
% Our policy should learn these trade-offs automatically; neither our reward nor our sampler is optimized to select cheaper client, because they may decrease performance if their datasets are highly imbalanced.
\commentblue{To enforce our budget constraint, the sampler ensures that the energy budget remains within the specified limit. If the energy were included in the reward function instead, this hard constraint could not be guaranteed.}

\subsection{Energy-Efficient RL Training}
\label{sec:training}
An important aspect of our energy-efficient framework is to guarantee that the RL training only consumes a minimal amount of energy.
As discussed in Section~\ref{sec:rwork}, the energy consumed by the RL training is not considered or evaluated in other RL-based FL client selection works.
The first step for our energy-efficient training is to make it independent of the number of clients, so that even with many clients, we can still train our RL model at runtime. % with only a minimal amount of energy.
One design choice that supports this independence is our MARL approach, where the state and action dimensions of each network are independent of the number of clients.
However, with an increasing number of FL clients, there will also be more actor/critic networks.
To keep the energy overhead constant, only the agents of clients who participated in an FL round update their policies; the others do not train.
This, however, leads to each updating agent exploring only one of the possible actions.
%To update the probabilities in our stochastic policy accordingly, it is important that we introduce a negative reward of comparable magnitude to the positive reward in order to adequately discourage non-participation and decrease the participation probability.
As discussed at the beginning of this Section~\ref{sec:bippo}, the negative reward is what allows the decrease of the participation probability for the clients that negatively impact the model performance, since non-participating agents do not update their policies.
The results of our evaluation (Section~\ref{sec:evaluation}) show that BIPPO is indeed able to learn a solid strategy, even if the models are only updated when the client participates.
Due to our training design choice, we can save 90\% of the energy used for RL training if only 10\% of clients are selected.
%A lower energy budget leads to a decreasing number of participating clients, which makes the energy used for RL training also decrease accordingly.
If more clients are included in the FL process, the training cost of BIPPO remains consistent.
More details on the energy consumed by our RL model are in Section~\ref{sec:overhead}.

\commentblue{It is important to note that not updating the policies of non-participating clients is not only beneficial in terms of energy. 
In BIPPO, the participation of a client does not only depend on the client's participation recommendation, but it is also highly dependent on the sampler (Section~\ref{sec:sampler}).
A ``good'' client that could improve the training accuracy might not be chosen in the first rounds, due to exploration.
It would therefore be forced into non-participation.
This could lead the client to mistakenly learn to decrease its own participation probability. 
To prevent this from happening, we only consider clients performing an action if they actually participate in the FL training.
}

\subsection{Sampling Strategies}
\label{sec:sampler}
Traditionally, in IPPO, the action decision is made for each actor network individually by sampling over the output probability distribution.
In our case, however, if not further controlled, this could lead to the FL process exceeding the global energy budget or using it inefficiently.
Given our energy budget, we therefore need a global sampler that selects as many clients as possible, without going over the budget.
We evaluate two sampling strategies for our global sampler: \textbf{stochastic policy sampling} and \textbf{$\epsilon$-greedy sampling}, inspired by how actions are chosen in PPO and DDQN, respectively.
Given the probabilities of the actor network, the stochastic action sampler of PPO randomly selects an action, where actions with higher probability are more likely to be chosen.
In contrast, the deterministic action selection strategy of DDQN chooses the action with the highest value. To guarantee exploration, DDQN needs an $\epsilon$ parameter that requires fine-tuning: whenever a random number is smaller than $\epsilon$, a random action is chosen. In each round, or every few rounds, $\epsilon$ decreases, progressively reducing exploration.
An overview of the client selection steps is provided in Algorithm \ref{alg:sampler}.
\begin{algorithm}
\caption{Budget-Based Client Selection for IPPO with different samplers}
\label{alg:sampler}
\begin{algorithmic}

\For{each agent $i = 1$ to $I$} 
\Comment{RL}
        \State Get action probability for client $i$: $a^t_i = \pi_i(s_i^t)$
\EndFor
\State cost $\gets 0$
\While{budget not reached} \Comment{Sampler}
    \If{policy = stochastic} \Comment{IPPO}
        \State Sample client $c \sim \mathbf{a}^t = \left[ a^t_i \right]_{i=1}^N$
    \ElsIf{policy = $\epsilon$-greedy} \Comment{BIPPO}
        \State Sample $e \sim \mathcal{U}(0,1)$
    \If{$e < \epsilon$}
        \State $c \gets$ random client
    \Else
        \State $c \gets \arg\max_i \; a^t_i$
    \EndIf
    \EndIf

    \If{cost + energy($c$) $<$ budget}
        \State $\text{cost} \gets \text{cost} + \text{energy}(c)$
        \State Add client $c$ to $\mathbf{\bar{a}}^t$
        \State Set $a^t_c \gets 0$
    \EndIf
\EndWhile
\State $\epsilon \gets \max(\epsilon \cdot 0.9, 0.05)$

\end{algorithmic}
\end{algorithm}

As shown in the algorithm, each agent $i$ first gets the probability ${a}_i^t$ for client $i$ to participate in the next FL round, given its policy $\pi_i$ and the current state $s^t_i$ of client $i$.
The stochastic policy randomly selects a client considering its probability, while the $\epsilon$-greedy sampler chooses a random client with probability $\epsilon$, otherwise it chooses the client with the highest probability.
After a client has been selected and added to the participating clients $\mathbf{\bar{a}}^t$, its probability is set to 0, and the process continues until the budget is reached.
At the end of one client selection round, the exploration for the $\epsilon$-greedy method is reduced by lowering $\epsilon$.

\begin{figure}[htbp]
  \centering
\includegraphics[width=\linewidth]{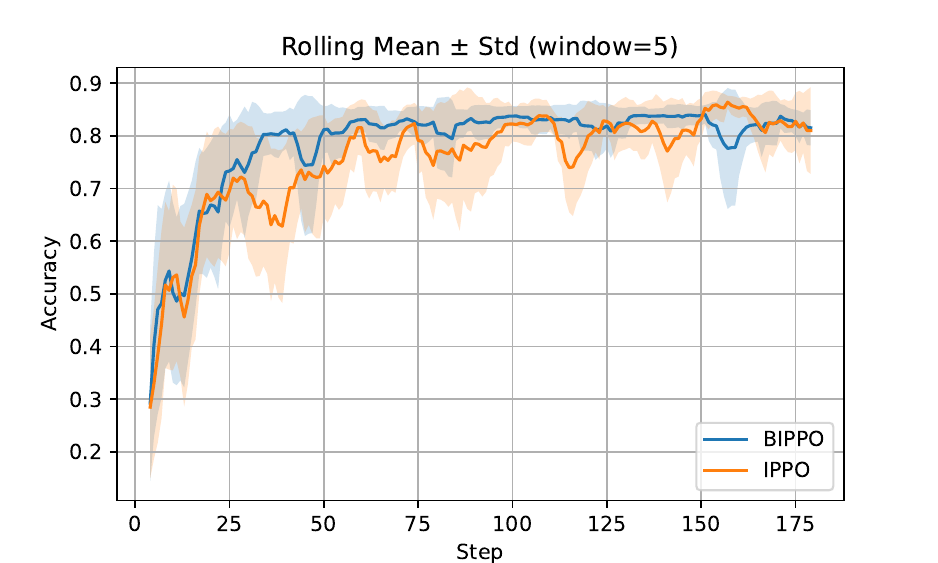}
    \caption{Comparison of different sampling strategies.}
    \label{fig:sampler}
\end{figure}

Figure \ref{fig:sampler} illustrates the performance of BIPPO with the different samplers. 
The data for this plot was collected by running FL on Fashion-MNIST with 20 clients. Further configuration details can be found in Section~\ref{sec:fashionMnist20}.
Figure~\ref{fig:sampler} shows that the $\epsilon$-greedy sampler used in BIPPO provides more consistent performance than the traditional stochastic sampler of IPPO.
Therefore, we adopt the $\epsilon$-greedy sampler instead of the traditional PPO sampling mechanism.
\commentblue{The $\epsilon$ decay parameter is set to 0.9, ensuring that the exploration at the beginning is high, while progressively reducing randomness as the training progresses.}

\subsection{Practical Considerations}
\label{sec:practical-considerations}
BIPPO requires that clients report their energy costs. The accuracy of the reported values depends on the measuring capabilities of the client devices. For example, for popular NVIDIA GPUs, the \texttt{nvidia-smi} utility or the \texttt{NVML} library\footnote{\url{https://docs.nvidia.com/deploy/nvml-api/index.html}} can be used to retrieve GPU power draw information during training. However, accurate monitoring of training energy costs is challenging~\cite{YangAA24}. 
Regarding the energy cost of communicating the computation results of an FL round, this is a function of the time a transmitter is active sending data (itself depending on link quality; the poorer the signal conditions, the longer the time to transmit the results), the size of the ML model, and the transmission power used. These vary depending on the network topology and the connectivity technology. When the latter applies transmit power control (e.g., 4G/5G), a conservative estimate can be derived assuming that devices transmit at the maximum power allowed. 
In the absence of precise energy measurement capabilities, devices can resort to models for estimating energy cost. This is also the approach we follow in this work, applying models, definitions, and assumptions typical in the FL literature~\cite{chen2024towards,zhang2022energyefficient,10552399}. We briefly elaborate on this approach in Section~\ref{sec:energy-model}.

%% file: sections/Experiments.tex
\section{Evaluation}
\label{sec:evaluation}
We evaluate BIPPO against competitive baselines, namely PPO, IPPO, our implementation of the FedPPO method, the FedAvg random selection approach, the Power-of-Choice method, and a custom variant of it, Power-of-Choice-Accuracy (PoCA). All baselines are detailed in Section~\ref{sec:baselines}. 
In this evaluation, we measure the models' accuracy to empirically assess the quality of the learned client-selection strategy under a fixed, constrained budget.
\subsection{Experimental Setup}

\begin{figure}[htbp]
    \centering

    % first image
    \begin{subfigure}[t]{0.48\linewidth}
        \centering\includegraphics[width=\linewidth]{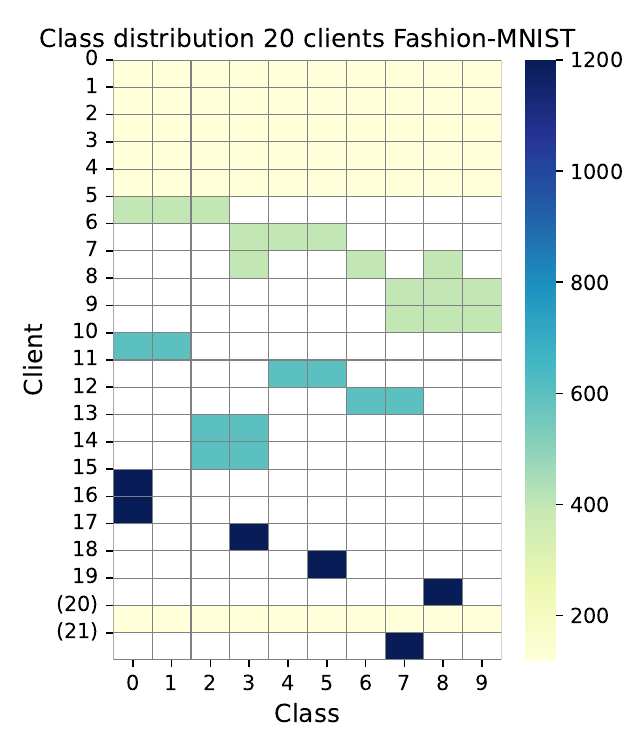}
        \subcaption{20 clients class distribution}
        \label{fig:fashion_mnist20}
    \end{subfigure}%
    \hfill
    % second image
    \begin{subfigure}[t]{0.48\linewidth}
        \centering\includegraphics[width=\linewidth]{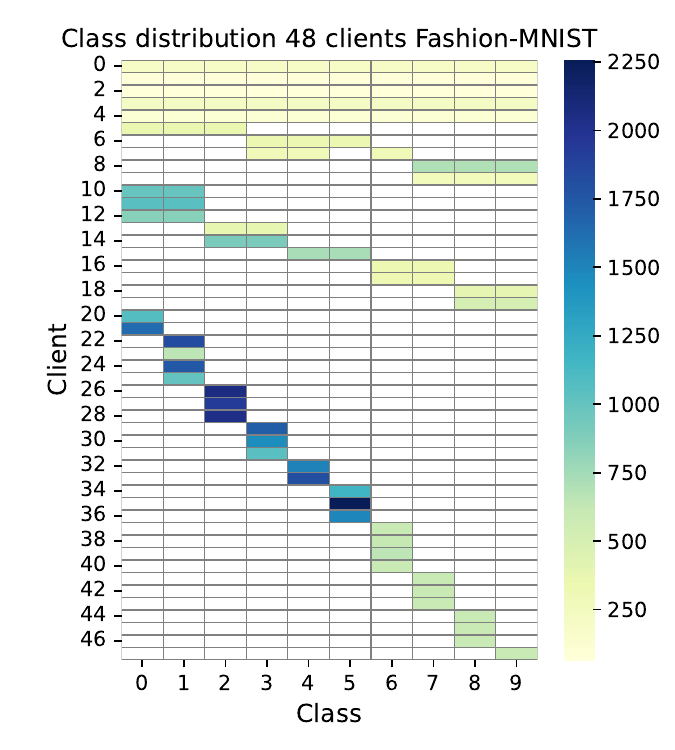}
        \subcaption{48 clients class distribution}
        \label{fig:48clients}
    \end{subfigure}
    
    \caption{Comparison of class distributions for different numbers of clients.}
    \label{fig:comparison}
\end{figure}

\noindent\textbf{Datasets}:
For our tests, we chose two classification datasets that are often used in FL tasks: Fashion-MNIST and CIFAR-10.
The non-IID label distribution for 20 clients is shown in Figure~\ref{fig:fashion_mnist20}. Clients fall into four categories: balanced clients hold samples of all labels, while the others hold samples of only three, two, or one label. With 20 clients, each category comprises a quarter of the clients. The label distribution was the same for Fashion-MNIST and CIFAR-10; the only difference is that, for CIFAR-10, the clients received fewer samples, as shown in Table \ref{tab:hyperparameters}.
In this figure, one can additionally see the two clients that join in the stability tests.
The class distribution for 48 clients is in Figure \ref{fig:48clients}.
Details on the models and the data size of each client are in Table \ref{tab:hyperparameters}.\\
% The clients in our experiments have four different types of datasets.
% Balanced clients have samples of all labels, some clients only have samples of three labels, other clients have samples of two labels, and extremely unbalanced clients only have one label. 
% For the tests with 20 clients, each of these categories is represented by a fourth of the clients.
\textbf{Hardware}: The tests conducted in this work ran on a server equipped with an AMD Ryzen 9 5950X processor (16 cores, 32 threads, maximum clock speed of 5.08 GHz) and two NVIDIA GeForce RTX 3090 GPUs with 24 GB of memory each.\\
\textbf{Hyperparameters}:
The hyperparameters for the FL process, the RL models, and the $\epsilon$-greedy sampler are listed in Table~\ref{tab:hyperparameters}.
% the README.md file in our Code~\footnote{\url{https://github.com/Lacki28/BIPPO/tree/main}}.
The energy parameters are taken from~\cite{liu2020client}, the model for Fashion-MNIST is taken from~\cite{vcilic2024reactive}, and the model of CIFAR-10 is taken from~\cite{he2016deep}.\\
\textbf{Budget}:
The budget is deliberately small, so that performance hinges on the right client selection. The overall budget is roughly one tenth of the total cost of all clients' training in one round --- 11\% for Fashion-MNIST and 10\% for CIFAR-10: an expensive client can be chosen only once per round, while two to three cheap clients fit within the budget.
\\
\textbf{GitHub}:
The GitHub repository contains the code to reproduce the results presented in the paper, including the run scripts used for the individual experiments.
Additionally, the repository includes a folder containing the results for each random seed.
For improved readability, the figures in the evaluation present the performance aggregated across all seeds, since the overall trends closely match those observed in the individual seed runs.
One notable exception is random seed ``7'', for which PPO exhibits substantially more consistent performance than for the other seeds. However, this behavior is not representative of the overall results.

% Fashion-MNIST model & Simple CNN \cite{vcilic2024reactive}\\
% Fashion-MNIST client training data &1.200\\
% Fashion-MNIST client test data &200\\
% CIFAR-10 model & Resnet-18 \cite{he2016deep}\\

\begin{table}[ht!]
\centering
\caption{FL, PPO, and $\epsilon$-greedy sampler Hyperparameters. The values within parentheses were only used in a subset of experiments.}
\label{tab:hyperparameters}
\begin{tabular}{|l c|}
\hline
\textbf{FL Hyperparameter} & \textbf{Value}  \\
\hline
Batch size          & 32\\
Local epochs        & 5\\
Learning rate        & 0.1\\
Fashion-MNIST model & Simple CNN \cite{vcilic2024reactive}\\
Fashion-MNIST client training data &1,200\\
Fashion-MNIST client test data &200\\
CIFAR-10 model & ResNet-18 \cite{he2016deep}\\
CIFAR-10 client training data &1,000\\
CIFAR-10 client test data &200\\

\hline
\hline
\textbf{Energy Hyperparameters \cite{liu2020client}} & \textbf{Value}  \\
\hline

Bandwidth ($B$) & $1 \ \text{MHz}$ \\
Transmitter power ($P_n$) & $0.5 \ \text{W}$ \\
Channel gain ($h_n$) & $10^{-8}$ \\
Background noise power ($\sigma$) & $10^{-10}$ \\
CPU cycles per bit & $20$ \\
$\tfrac{\alpha_i}{2}$ & $2 \times 10^{-28}$ \  \\
CPU-cycle frequency ($f$) & $700;(1,430);1,500 \ \text{MHz}$ \\

\hline
\hline
\textbf{PPO Hyperparameter} & \textbf{Value} \\
\hline
Batch size          & 2\\
Local epochs        & 8\\
Hidden layer size   & 128\\
Learning rate       & 0.001 \\
Discount factor ($\gamma$)        & 0.9\\
Clip parameter $\epsilon_{\text{clip}}$                    & 0.2 \\
Bias-variance tradeoff ($\mu$)    & 0.8 \\
\commentblue{Seeds} & \commentblue{7;28;42;(231)}\\
\hline
\hline
\textbf{$\epsilon$-greedy Hyperparameter} & \textbf{Value} \\
\hline
Epsilon ($\epsilon$)              & 1.0\\
Epsilon decay                     & 0.9\\
Minimum epsilon ($\epsilon_{\min}$) & 0.05 \\
\hline
\end{tabular}
\end{table}

\subsection{Baselines}
\label{sec:baselines}
To compare our method, we consider various heuristic and RL baselines:\\
\textbf{FedAvg (Heuristic)}: The sampler of this approach randomly selects the clients until the budget is reached~\cite{zhao2025fedppo, mcmahan2017communication}.\\
\textbf{PoC (Heuristic)}: The \textit{Power-of-Choice} method~\cite{cho2020client} selects the clients with the highest loss until the budget is reached.
Using clients with the highest loss often improves model convergence, as shown by~\cite{cho2020client}.
\\
\textbf{PoCA (Heuristic)}: The \textit{Power-of-Choice-Accuracy} method is our adaptation and implementation of the Power-of-Choice method; instead of selecting the clients with the highest loss, it selects the clients with the lowest accuracy until the budget is reached.
The lowest-performing clients likely hold data that is underrepresented in the current global model.
\\
\textbf{FPPO (RL)}: FPPO is our implementation of FedPPO~\cite{zhao2025fedppo}, adapted to our use case. Our FPPO computes the participation suggestions for all clients through a single network and selects the clients with the highest values until the budget is reached. To train the policy, it computes the reward and stores the samples of the participating clients. \\ 
\textbf{PPO (RL)} We implement PPO with a traditional stochastic-policy sampler, i.e., a weighted random selection within the budget boundaries. \\
\textbf{IPPO (RL)} This baseline has the same sampler as PPO, but instead of a single network with a global view, each client has its own network.

\subsection{Non-IID and Heterogeneous Device Performance}
\label{sec:fashionMnist20}
%These tests evaluate the influence of the different client selection methods on the overall FL model's performance. 
We evaluate our approach in a setting with 20 clients using the Fashion-MNIST dataset.
The data distribution is non-IID, as shown in Figure~\ref{fig:fashion_mnist20}, the clients have different device types, and the goal is to select clients that optimize performance in an energy-constrained environment.
To simulate heterogeneous devices, we included 8 \textit{cheap} (low-energy) and 12 \textit{expensive} (high-energy) clients, distinguished by the CPU-cycle frequency in Table \ref{tab:hyperparameters}.
%The budget for the participating clients in each round is set to approximately 11\% of the total energy that would be consumed if all clients were training for one round.
Relative to the selected budget (i.e., 11\% of the total), a cheap client takes about 33\% of this budget to train, and thus up to 3 clients of this type could be selected.
Only one expensive client can be selected in each round, since each of them needs about 56\% of this budget.
\begin{figure}[htbp]
  \centering
    \includegraphics[width=\linewidth]{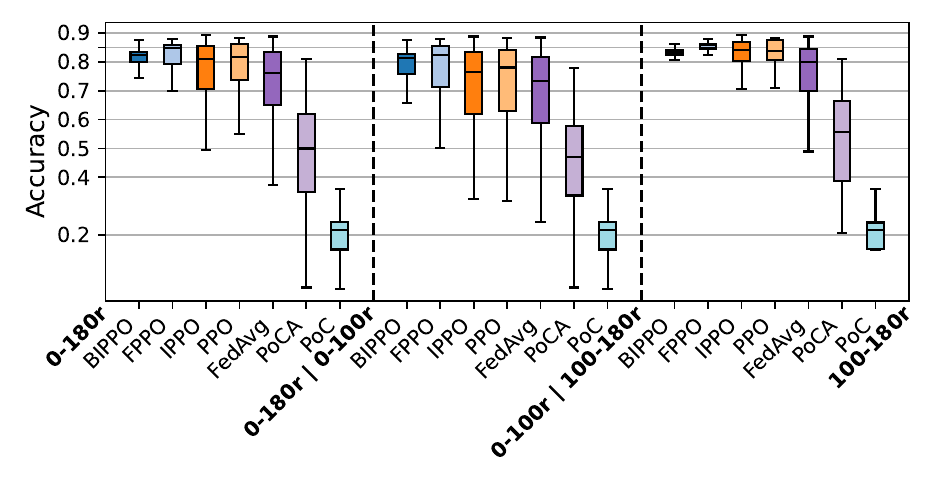}
    \caption{11\% Energy budget, 20 clients training on Fashion-MNIST.}
    \label{fig:acc_budget1}
  
\end{figure}

Looking at the results of Figure \ref{fig:acc_budget1}, one can see that BIPPO and FPPO have the best performance, since they achieve high accuracy and have a low interquartile range (IQR), which indicates consistent performance.
The RL methods with the traditional stochastic sampler (IPPO, PPO) also achieve high performance, but with a higher IQR, indicating less consistent global performance.
The heuristic approaches (FedAvg, PoCA, PoC) have lower performance; the difference is especially visible in the last rounds.
PoC performed so badly in these first tests that we introduced PoCA, which we will use for the other evaluation sections.

\subsection{Generalizability}
\label{sec:cifar10}
To assess how well our results \textit{generalize}, we test our method on a more challenging task, training a ResNet-18 on CIFAR-10. Here, with generalizability, we mean that our approach, trained from scratch, performs well in a different FL setting, i.e., a new dataset, model, and energy profile, without reusing any previously learned policy.

As in the first tests, we use 8 cheap and 12 expensive clients.
For the CIFAR-10 dataset, we set our energy budget to about 10\%, slightly below the 11\% used in the Fashion-MNIST tests. The budget ratio for cheap and expensive clients is changed to 47\% and 52\%, respectively.

\begin{figure}[htbp]
  \centering
\includegraphics[width=\linewidth]{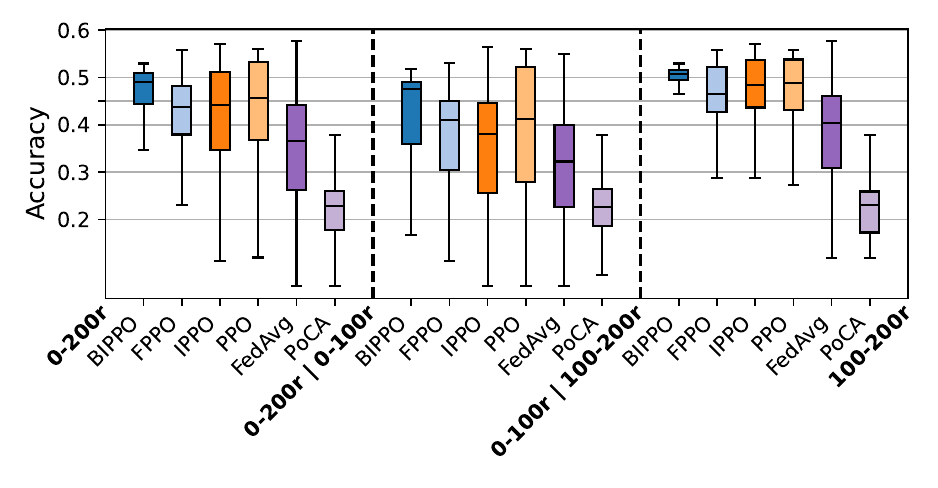}
    \caption{Accuracy distribution on CIFAR-10.}
    \label{fig:stepCifar10}
\end{figure}

The results of the CIFAR-10 tests, visualized in Figure \ref{fig:stepCifar10}, show that BIPPO outperforms all other methods, with the highest median and the smallest IQR, indicating consistently high performance.
The other RL methods perform comparably to one another, with FPPO showing a slightly lower IQR.
Both heuristic methods perform worse than the RL methods.

\subsection{Transfer Learning}
\label{sec:transfer}
Different from Section~\ref{sec:cifar10}, in this section we analyze the transfer learning performance. Concretely, we want to inspect whether strategies learned in one setting can be reused in another to provide a jump-start, without full retraining.
Most existing contributions perform extensive RL training directly on the target environment.
The main advantage of pretraining is to provide a turnkey model at runtime, or at least a warm start, so the strategy can be as fast as possible while still achieving close to optimal performance. However, time and data-collection constraints make it difficult to achieve deep, reliable training that does not overfit, and the training stage itself incurs computational and energy-related costs.
% ADVANTAGE: Accurate model - Warm start, closer to optimal performance
% DISADVANTAGE: Time/authorization required to collect accurate data + computational & energetical training cost + overfitting to a single environment

%To make our approach relevant in settings where training is inaccessible or counterproductive, we evaluate the convergence and energy efficiency of BIPPO.
%Since one usually does not have access to data for the exact same environment, we use the data collected from the 
\commentblue{Based on the ``learning from easy missions'' principle~\cite{taylor2009transferlearning}, we transfer the RL models pretrained on Fashion-MNIST --- each agent's actor and critic networks --- to an FL process aimed at classifying the CIFAR-10 data. This strategy is one of the possible approaches for transfer learning~\cite{taylor2009transferlearning}. 
%The rationale is to transfer knowledge from the source task(s) to the target task, e.g., to help jump-start the model's performance. 
In particular, transfer learning is useful when the state is large or continuous, as learning these problems ``tabula rasa'' is cumbersome~\cite{taylor2009transferlearning}.}
Specifically for this use case, the CIFAR-10 images are more difficult to predict than the Fashion-MNIST ones. %, requiring greater training complexity, higher costs, and a larger budget.
%To increase the shift from the Fashion-MNIST use case
Concretely, we change the classification model to a larger network, which in turn raises the clients' absolute energy consumption and requires rescaling the overall energy budget --- under the old budget, no client could be selected. The transfer is well-posed because the client population is unchanged: the same number of clients, the same cheap/expensive device mapping, and the same per-client data distributions; only the absolute energy values change.
Furthermore, as a benchmark, we consider IPPO, pretrained on Fashion-MNIST.
%It should be noted that the two setups differ in terms of dataset, FL model type, budget, and energy consumption, but the data distribution and the distribution of cheaper and more expensive devices were the same. Due to a comparable distribution of data and energy consumption, the previously learned policies can be applied to this new setting.

\begin{figure}[htbp]
  \centering
\includegraphics[width=\linewidth]{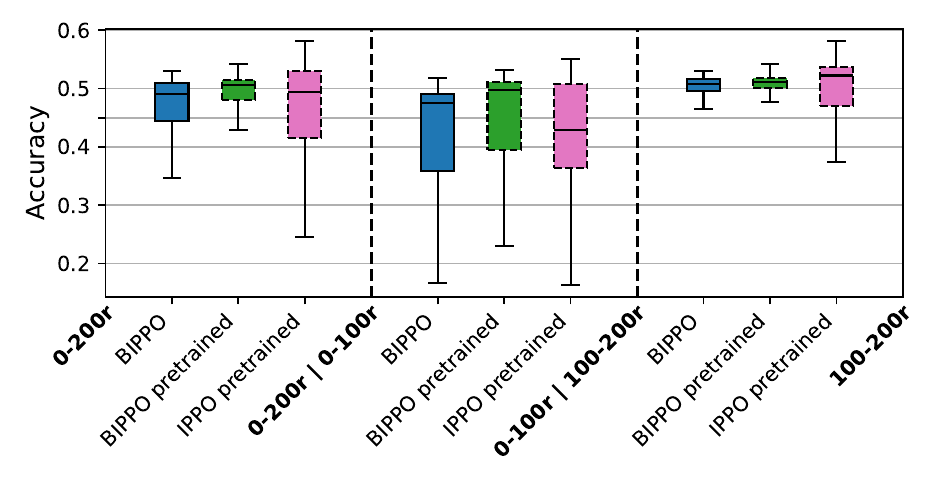}
    \caption{Accuracy distribution of CIFAR-10 pretrained.}
    \label{fig:pretrain1}
\end{figure}
\begin{figure}[htbp]
  \centering
\includegraphics[width=\linewidth]{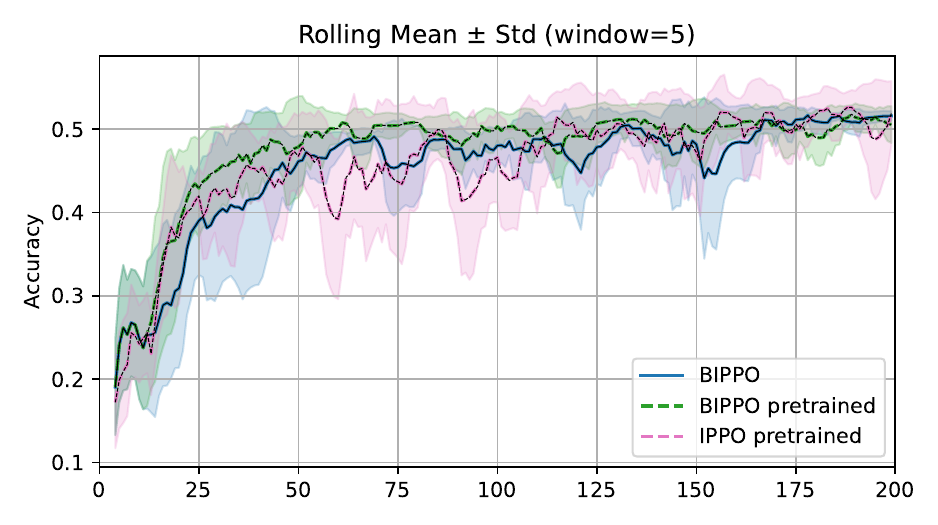}
    \caption{CIFAR-10 pretrained, rolling mean $\pm$ standard deviation.}
    \label{fig:pretrain2}
\end{figure}

%In this experiment, we evaluate the jump-start gain that transfer learning gives to BIPPO, as well as the performance benefit over IPPO.
Figure \ref{fig:pretrain1} shows the aggregated results. 
The value of transfer learning is immediately visible, with solid performance for the pretrained IPPO and, even more prominently, the pretrained BIPPO. Pretraining yields a clear jump-start on CIFAR-10: the pretrained BIPPO achieves an almost 10\% accuracy gain over the non-pretrained BIPPO, as Figure~\ref{fig:pretrain2} shows.

%take
% BIPPO (ϵ-greedy)
% 0.44530805348258706
% BIPPO (stochastic policy)
% 0.40452886090381424
% BIPPO pretrained (ϵ-greedy)
% 0.46878973535102264
% BIPPO pretrained (stochastic policy)
% 0.44742727508291874

% BIPPO (ϵ-greedy)
% 0.48543300295398006
% BIPPO (stochastic policy)
% 0.46203168921019905
% BIPPO pretrained (ϵ-greedy)
% 0.5000627487562189
% BIPPO pretrained (stochastic policy)
% 0.4946346470771144

\subsection{Stability under Client Churn}
\label{sec:dynamic}
%In a practical use case, IoT devices may join or leave the FL process over time.
%Therefore, it is important for an RL method to be robust to such changes.
%For the following tests, we do not focus on transferring previously learned knowledge to a new policy.
%Instead, 
The purpose of this evaluation is to empirically verify how different RL architectures (SARL, MARL) perform in dynamic environments, i.e., when FL clients join or leave the active pool (client churn). The magnitude of the required adaptation depends on the architecture.
In MARL approaches, such as BIPPO and IPPO, it is sufficient to add or remove the actor and critic networks for joining/leaving clients, whereas in centralized SARL approaches, such as PPO and FPPO, we need to initialize a new network. 

In the experiment, we let two new clients join the FL process after 100 rounds: a cheap one with a balanced data distribution and an expensive one with an unbalanced distribution.
After 150 rounds, two clients leave (10\% of the original). We deliberately pick two of the first five clients, i.e., the ones with a balanced data distribution.

\begin{figure}[htbp]
  \centering
\includegraphics[width=\linewidth]{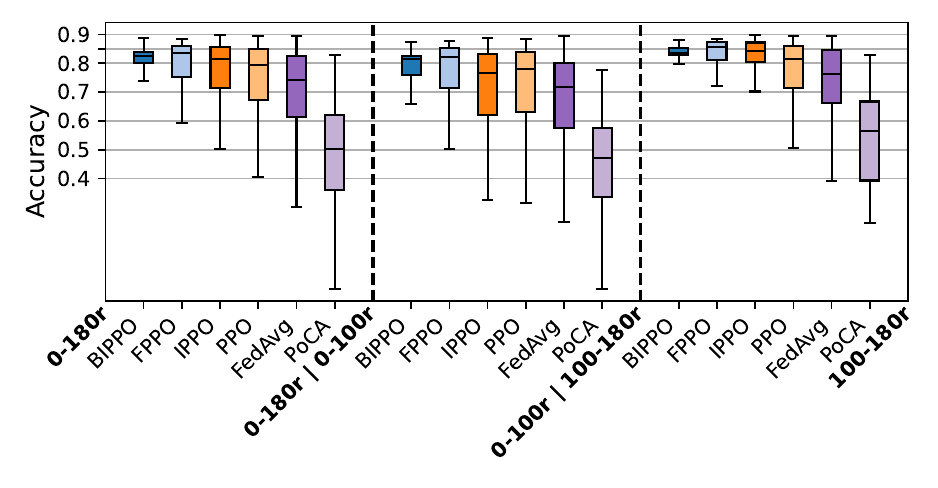}
    \caption{Accuracy distribution in dynamic environments.}
    \label{fig:dynamic}
\end{figure}

When comparing rounds 100-180, in which some clients join and leave, the performance difference between the two MARL approaches, BIPPO and IPPO, is limited, as shown in Figures \ref{fig:acc_budget1} and \ref{fig:dynamic}.
As hypothesized, the SARL PPO and FPPO models experience significant performance drops. The result leads us to conclude that the main cause is the retraining required as the state and action spaces change, consistent with the observations in Section~\ref{sec:transfer}.
Considering the classical FL client selection methods, only FedAvg's performance slightly decreases, compared to Figure \ref{fig:acc_budget1}.

\begin{figure}[htbp]
  \centering
\includegraphics[width=\linewidth]{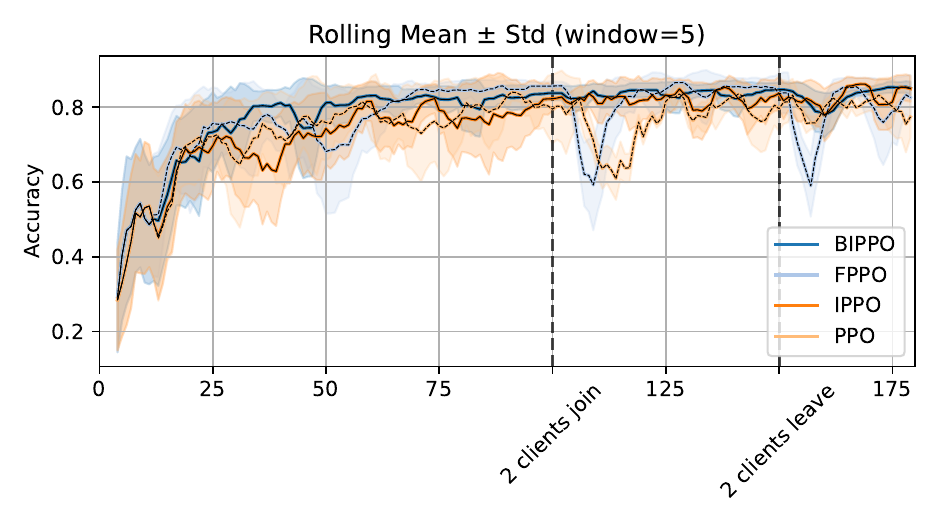}
    \caption{Rolling mean $\pm$ standard deviation of RL methods in dynamic environments.}
    \label{fig:dynamic_rolling}
\end{figure}
%BIPPO (ϵ-greedy)
% 0.8339392476369506
% PPO (ϵ-greedy)
% 0.8031977367289634
% BIPPO (stochastic policy)
% 0.8242305674010899
% PPO (stochastic policy)
% 0.77728214410227
% Random
% 0.7420111677560894
% Highest Loss
% 0.5434003542891602

A clearer perspective is provided by the rolling mean progression displayed in Figure~\ref{fig:dynamic_rolling}.\footnote{For the sake of clarity, we omit the line plots of the heuristic baselines, as their performance is much lower.}
The performance degradation of PPO and FPPO when new clients join is clearly visible, especially if compared with BIPPO's and IPPO's behaviors.
When clients leave, all methods remain stable, except for FPPO, whose accuracy significantly drops. This drop can be explained by FPPO's sampler having preferred a client that has left: until the value estimates adapt, the process selects clients that lead to worse performance.
% After this client leaves, the process may choose a client that leads to worse performance.
% Due to the performance degradation, the reward will be negative, and the probability of choosing this client will decrease.
% After this probability decreased, there may be another client with a higher probability that the sampler will then prefer, leading to better accuracy again.
%Overall, BIPPO's design proves its robustness, even in dynamic environments.

\subsection{Scalability}
\label{sec:48}
These tests address the \textit{performance} dimension of scalability. We check whether FL accuracy remains consistent as the client population grows. We run experiments with 48 clients, more than double the number in the previous settings, and within the limit of clients our hardware supports.
To make the environment even more challenging, clients in this setting have three different device types, non-IID data, and varying data sizes.
We ran the experiments for 300 rounds.

\begin{figure}[htbp]
  \centering
    \begin{subfigure}{0.8\linewidth}
    \centering
    \includegraphics[width=\linewidth]{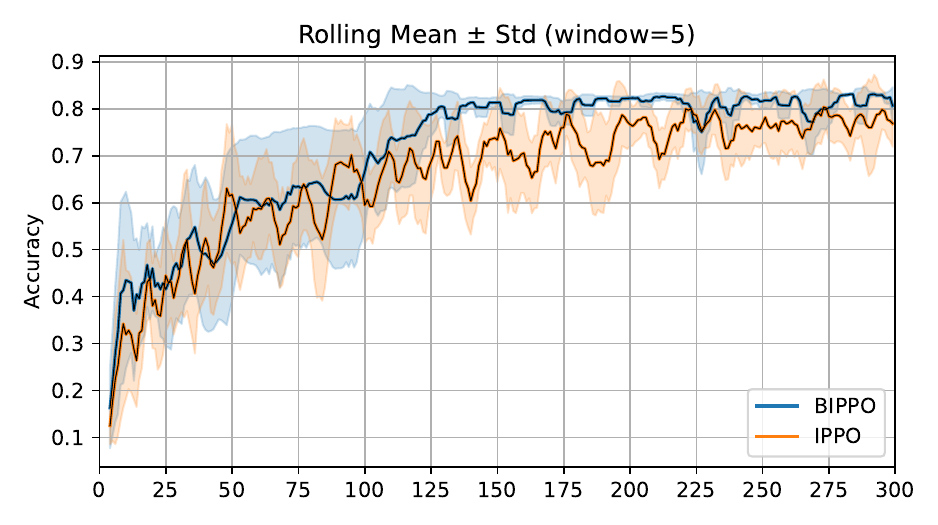}
    \caption{Rolling mean $\pm$ standard deviation of BIPPO and IPPO.}
    \label{fig:ippo}
  \end{subfigure}
  \hfill
  \begin{subfigure}{0.8\linewidth}
    \centering
    \includegraphics[width=\linewidth]{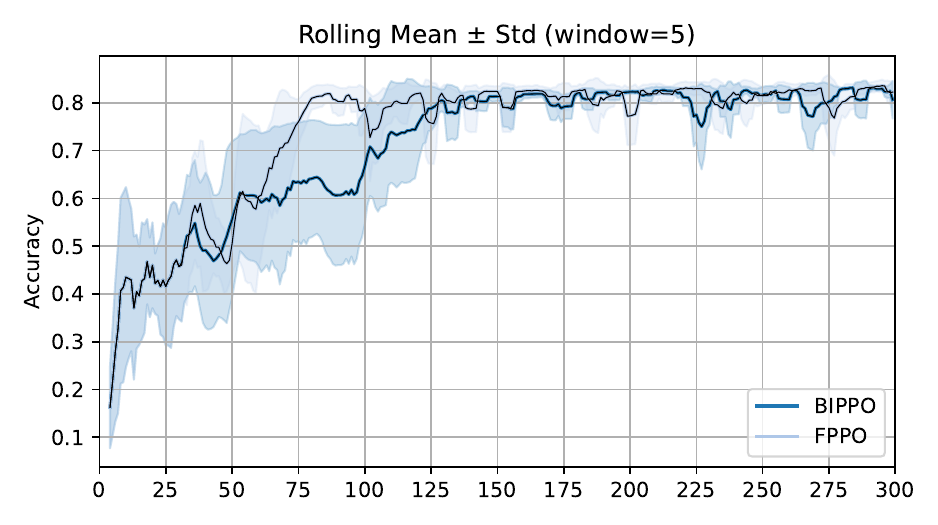}
    \caption{Rolling mean $\pm$ standard deviation of BIPPO and FPPO.}
    \label{fig:fppo}
  \end{subfigure}
  \caption{Comparison of BIPPO scalability results.}
  \label{fig:fppo48cl}
\end{figure}

Our larger-scale tests show that BIPPO outperforms IPPO even in this larger setting, as illustrated in Figure \ref{fig:ippo}.
Compared to FPPO, it takes longer to converge, as shown in Figure \ref{fig:fppo}.
While FPPO takes about 75 rounds to reach a global accuracy of about 80\%, BIPPO needs about 125 rounds.
FPPO may find a good selection policy faster, but the energy cost of training FPPO in larger settings increases compared to BIPPO, which maintains consistent energy consumption across settings, as discussed in Section \ref{sec:overhead}.
\begin{figure}[htbp]
  \centering
\includegraphics[width=\linewidth]{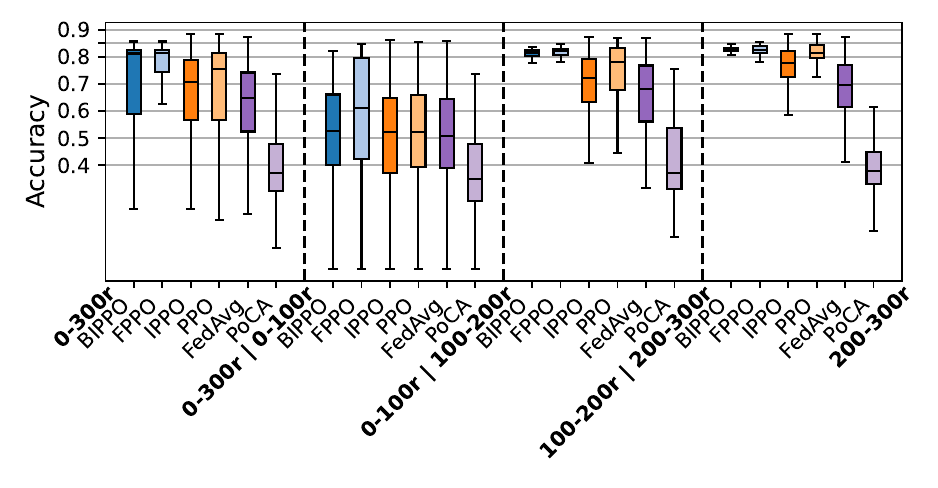}
    \caption{Performance with 48 clients training on Fashion-MNIST.}
    \label{fig:48cl}
\end{figure}
Figure \ref{fig:48cl} shows that our RL methods again outperform heuristic approaches, with BIPPO and FPPO reaching higher accuracy than PPO and IPPO.

\subsection{Sustainability}
\label{sec:overhead}
As argued in Section~\ref{sec:intro}, a scalable system must also be a sustainable one. Hence, we provide a comparison of the additional computational load (and, in turn, energy cost) generated by the RL methods. 
For the energy consumption comparison, we used the MAC (Multiply-Accumulate) metric to count operations. 
%\cite{yang2017method}.
\footnote{An approximate translation from FLOPs to MACs can be achieved by dividing the number of FLOPs by two \cite{getzner2023accuracy}.}
%In detail, to obtain a fully accurate energy estimation, we should multiply MACs by the buffer size and by the number of epochs. However, as these last two terms are invariant in SARL and MARL, we can cancel them out.
In summary, while not an exact measure of energy consumption~\cite{yang2017method}, MACs are a sufficient metric for estimating the relative overhead of the RL methods with respect to the FL training.

\begin{figure}[htbp]
  \centering
    % \begin{subfigure}{0.48\linewidth}
    \centering
    \includegraphics[width=0.8\linewidth]{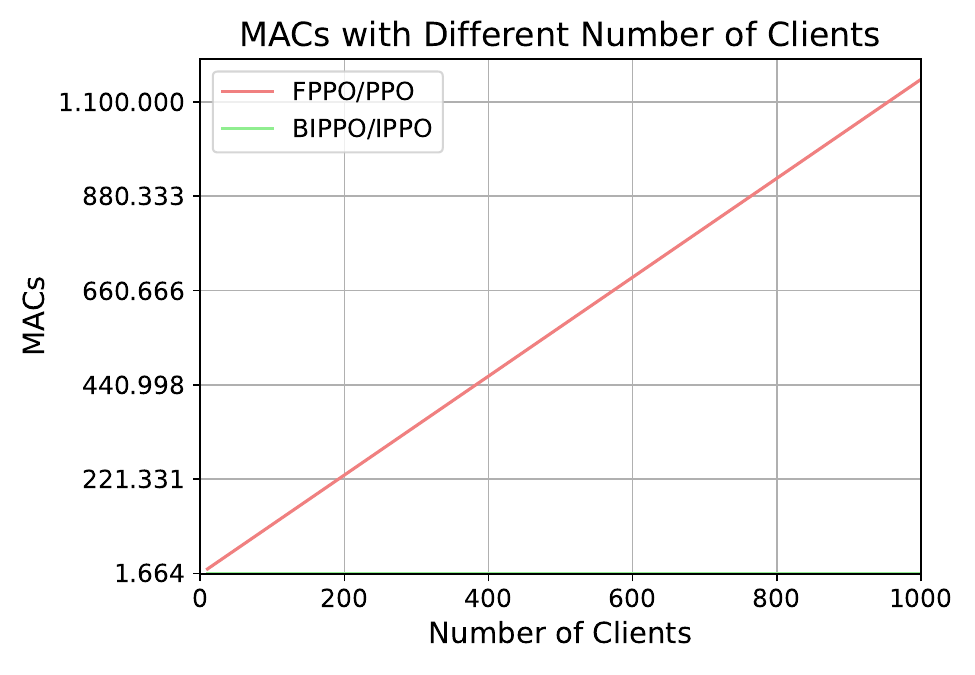}
  %   \caption{MACs comparison between SARL and MARL.}
  %   \label{fig:macs}
  % \end{subfigure}
  % \hfill
  % \begin{subfigure}{0.48\linewidth}
  %   \centering
  %   \centering
  %   \includegraphics[width=\linewidth]{figures/time_complexity2.pdf}
  %   \caption{Inference time comparison SARL and MARL.}
  %   \label{fig:time_compexity}
  % \end{subfigure}
  \caption{MACs comparison between SARL and MARL.}
  \label{fig:sustainability}
\end{figure}

Figure~\ref{fig:sustainability} provides an overview of the MACs needed for the NN to perform a full pass on the dataset, considering up to 1,000 clients.
For these measurements, we used torchprofile.\footnote{\url{https://github.com/zhijian-liu/torchprofile}}
With 20 clients, SARL needs 23,424 MACs for one NN pass, roughly 14 times the 1,664 MACs BIPPO needs.
In the larger 48-client setting, SARL needs 55,680 MACs, more than double the 20-client setting, while for BIPPO, it remains constant. Due to its multi-agent design, BIPPO allows the parallelization of both action selection and training, reducing computation time and improving the system's scalability.
In detail, action selection requires one pass per agent in MARL and is therefore linear in the number of clients; yet, BIPPO's models remain small and constant in size, making this step lightweight. The training step is where BIPPO really shines: both SARL and MARL approaches perform the same number of updates per round, one per participating client, but each BIPPO update passes a small, constant-size network, whereas each SARL update passes a network whose size grows with the client population. This makes BIPPO's training energy independent of the total number of clients.

Taking another viewpoint for energy consumption, i.e., the impact of the ML model and data size, we can put BIPPO into perspective. One image in the CIFAR-10 dataset requires 556,037,632 MACs, whereas in Fashion-MNIST, the number is 2,751,000 MACs.
To get the exact number of MACs for one training client, we should multiply the MACs per image by the number of samples and the number of epochs. Furthermore, the larger the budget, the more samples are used for training, and the greater the energy consumption.
Nonetheless, given that the RL-related MACs are up to five orders of magnitude lower, we can say that the RL methods only add minimal additional energy impact, making the solution appropriate for resource-constrained real-world scenarios.

%% file: sections/Discussion.tex
\section{Discussion}

\begin{table*}[ht!]
\centering
\caption{Summary of the experiments; the best result is in bold, the second best is underlined.}
\label{tab:summary_exp}
\begin{tabular}{|l l l | c | c c c c c|}
\hline
\textbf{Section} & \textbf{Challenge} & \textbf{Metric} & \textbf{BIPPO}& \textbf{FPPO}& \textbf{IPPO} & \textbf{PPO} & \textbf{FedAvg}& \textbf{PoCA}\\
\hline

%FashionMNIST
\ref{sec:fashionMnist20}& non-IID & rounds to target (mean) & \underline{43.0}& \textbf{41.75} & 66.75 & \underline{43.0} & 88.25 & - \\
\ref{sec:fashionMnist20}& non-IID & rounds to target (std) & \textbf{10.416}& 13.663 & 17.297 & \underline{12.748} & 41.342 & - \\
\ref{sec:fashionMnist20}& non-IID & mean accuracy & \underline{0.779} & \textbf{0.787} &0.751 &0.759 &0.701 &0.491\\
\ref{sec:fashionMnist20}& non-IID & std of accuracy  & \textbf{0.14}& \underline{0.147} &0.157 &0.157 &0.158 &0.163\\

\ref{sec:cifar10}& Generalizability &rounds to target (mean)  & \textbf{36.25} &\underline{38.0} & 42.0 & 44.5 & 80.25 & - \\
\ref{sec:cifar10}& Generalizability &rounds to target (std)  & \underline{9.808} &\textbf{9.67} & 13.019 & 16.286 & 12.577 & - \\
\ref{sec:cifar10}& Generalizability & mean accuracy & \textbf{0.452}& 0.415 &0.413 &\underline{0.427} &0.352 &0.226\\
\ref{sec:cifar10}& Generalizability &  std of accuracy & \underline{0.096}& 0.105 &0.121 &0.121 &0.114 &\textbf{0.053}\\

\rowcolor{gray!25}
\ref{sec:dynamic}& Stability & rounds to target (mean) & \underline{43.0}& \textbf{41.75} & 66.75 & \underline{43.0} & 89.5 & - \\
\rowcolor{gray!25}
\ref{sec:dynamic}& Stability& rounds to target (std) & \textbf{10.416}& 13.663 & 17.297 & \underline{12.748} & 43.396 & - \\
\rowcolor{gray!25}
\ref{sec:dynamic}& Stability& post-target accuracy (mean) &\textbf{0.784} &\underline{0.767} &0.754 & 0.739& 0.698&0.495\\
\rowcolor{gray!25}
\ref{sec:dynamic}& Stability& post-target accuracy (std) &\textbf{0.139} & \underline{0.158} & \underline{0.158}&\underline{0.158}& 0.16&0.16\\

\ref{sec:48}& Scalability& rounds to target (mean) & \underline{110.333}& \textbf{72.667} & 201.333 & 197.333 & 242.5 & - \\
\ref{sec:48}& Scalability &rounds to target (std) & 32.014& \textbf{9.843} & \underline{22.455} & 54.199 & 57.5 & - \\
\ref{sec:48}& Scalability &mean accuracy & \underline{0.709} & \textbf{0.736} &0.657 &0.678 &0.621 & 0.396 \\
\ref{sec:48}& Scalability &std of accuracy &0.178 & \underline{0.17} &0.178 &0.183&\textbf{0.164} & 0.133\\

\rowcolor{gray!25}
\ref{sec:overhead} & Scalability & RL NN MACs 20 clients &\textbf{1,664}&\underline{23,424}&\textbf{1,664}&\underline{23,424}&-&-\\
\rowcolor{gray!25}
\ref{sec:overhead} &Scalability & RL NN MACs 48 clients & \textbf{1,664} &\underline{55,680}& \textbf{1,664}&\underline{55,680}&-&-\\
\rowcolor{gray!25}
% \ref{sec:overhead} &Scalability & RL NN Inference time (ms) 46 clients & \underline{0.052} &0.055& \underline{0.052}&0.055&\textbf{0}&\textbf{0}\\
% \rowcolor{gray!25}
% \ref{sec:overhead} &Scalability & RL NN Inference time (ms) 1000 clients & \underline{0.052} &0.960& \underline{0.052}&0.960&\textbf{0}&\textbf{0}\\
\hline
\end{tabular}
\end{table*}

Table \ref{tab:summary_exp} summarizes the main results of the evaluation.
A few rows are highlighted in grey. These rows are emphasized as these are metrics that, to the best of our knowledge, no other RL-based FL client selection work has considered.
This table reports the \textit{rounds to target}: the round at which the average accuracy of the last 10 rounds reaches the target accuracy~\cite{zhao2025fedppo}. For Fashion-MNIST, the target accuracy is 80\%, taken from \cite{wu2021fast}, while for CIFAR-10 it is 40\%.
The average accuracy reached with CIFAR-10, given the budget and the non-IID environment, was between 35\% for FedAvg and 45\% for BIPPO.
We therefore decided on the target accuracy of 40\% as a representative mid-point.

The first finding from this table is that heuristic methods may consume less energy, but their performance falls short of the RL methods.
The traditional PPO methods that use the stochastic sampler perform worse than the RL methods that leverage clients' ranking (BIPPO's $\epsilon$-greedy and FPPO's greedy sampler).
Table \ref{tab:summary_exp} shows that, in terms of mean accuracy or mean rounds to target, BIPPO always performs best or second best.
In summary, BIPPO demonstrates good performance while consuming only a minimal amount of additional energy, which is independent of the total number of FL clients. Importantly, it is also robust to client churn, maintaining good performance in dynamic environments.

\noindent\textbf{Limitations.} The energy consumption in this work was estimated using a formula, which is a standard approach in related studies that facilitates comparison to existing work.
Furthermore, the hyperparameters to calculate the energy consumed in each FL round were also taken from previous studies.
\commentblue{To compare the energy used for training the RL methods, we used MACs, which provide a good relative measure, but they do not allow for a precise mapping to energy consumption. This is because data movement, rather than computation, dominates energy consumption, and because the memory hierarchy and dataflow have a large impact on data-movement energy consumption~\cite{yang2017method}.
However, energy measurements are challenging and can be inaccurate. Parameters such as the sampling frequencies can highly influence measured energy consumption~\cite{Jagannadharao2024ABG}. 
Therefore, using MACs is a sufficient metric for relative comparison, and in the future, we aim to measure and compare actual energy consumption.}
The datasets in this work are commonly used in related FL research, ensuring comparability.

\noindent\textbf{Future work.} To improve convergence in larger settings, we want to optimize transfer learning to further improve the learned strategy. We aim to pretrain a model that will generalize well over various settings to help MARL converge faster.
Additionally, we want to consider security aspects, particularly in scenarios where clients may behave maliciously. Finally, testing our approach on a dataset for a different learning task is an interesting future direction.

\section{Conclusion}
\label{sec:conclusion}

This paper presents BIPPO, an RL-based method for selecting IoT devices to participate in an FL task within a limited energy budget, making it suitable for real-world scenarios.
BIPPO quickly and autonomously learns a good strategy that outperforms conventional heuristic client selection methods and traditional RL methods in various environments, achieving higher mean accuracy.
Compared to the SARL approach FPPO, BIPPO achieves comparable or better accuracy in smaller settings, although it needs more rounds in larger settings to find a good selection strategy.
However, FPPO's performance comes at a much higher cost: its training consumes more than 14 times BIPPO's energy already in smaller settings, and this gap widens with the number of clients (33$\times$ at 48), while BIPPO's overhead remains constant.
Importantly, compared to SARL, BIPPO also allows seamless integration of new clients without the need to retrain the entire system from scratch. The evaluation showed that BIPPO indeed maintains its performance in dynamic environments where clients join and leave.

\section*{Acknowledgments} This work has been supported by the European Union's Horizon Europe under grant agreements No. 101079214 (AIoTwin) and No. 101135576 (INTEND).
Thanks to Axel Brunnbauer for the valuable help and expert feedback on RL and PPO.

%% file: sections/bibliographies.tex
% \section*{Biographies}

\begin{IEEEbiography}[{\includegraphics[width=0.8in, keepaspectratio]{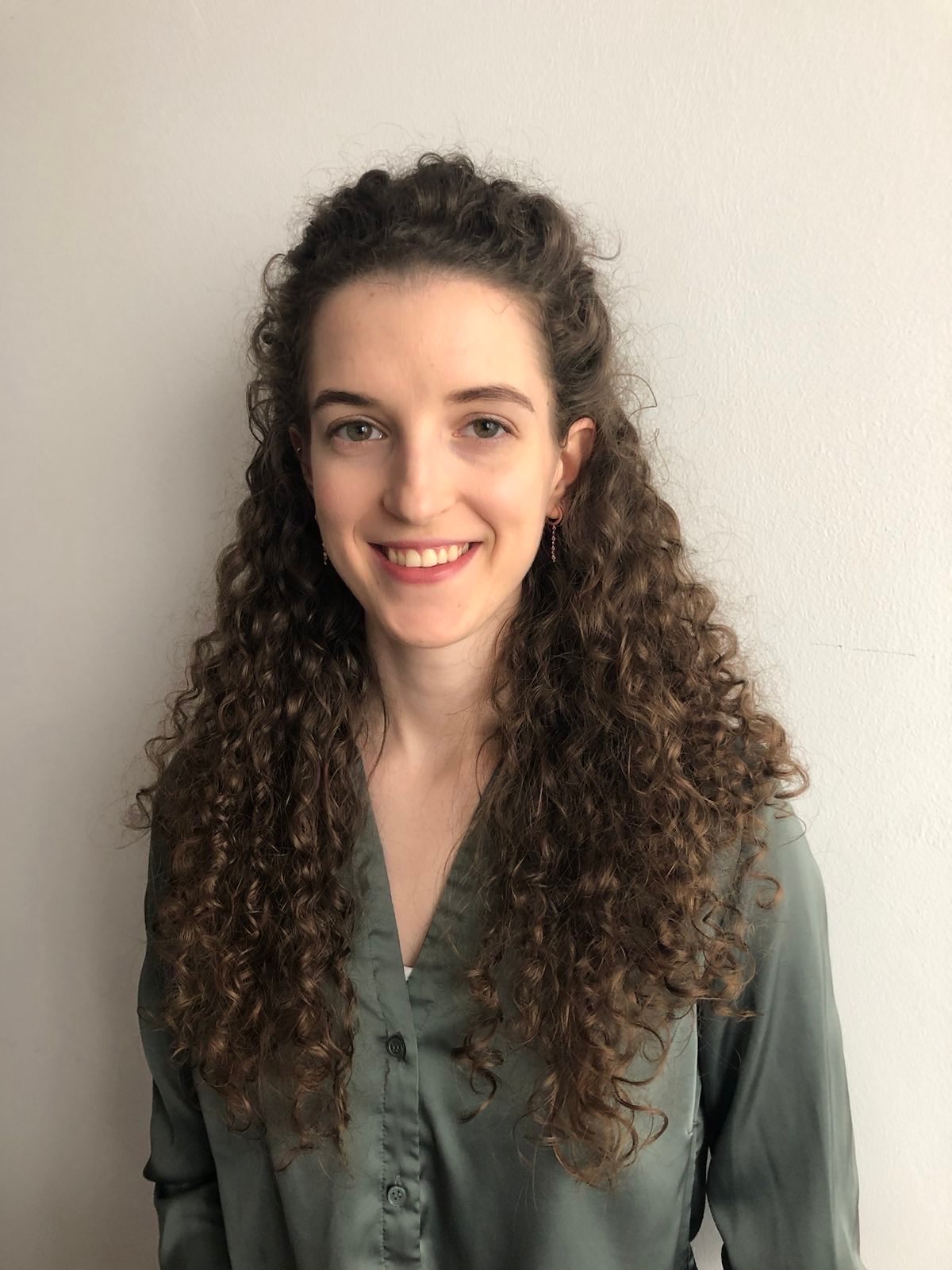}}]{Anna Lackinger} received the MSc degree from the Technical University of Vienna, Austria, in 2023, with distinction in the field of computer science. She is now working toward a PhD degree in the Distributed Systems Group at the Technical University of Vienna in Austria. Her research interests include edge intelligence and machine learning.
\end{IEEEbiography}

\begin{IEEEbiography}[{\includegraphics[width=0.8in,keepaspectratio]{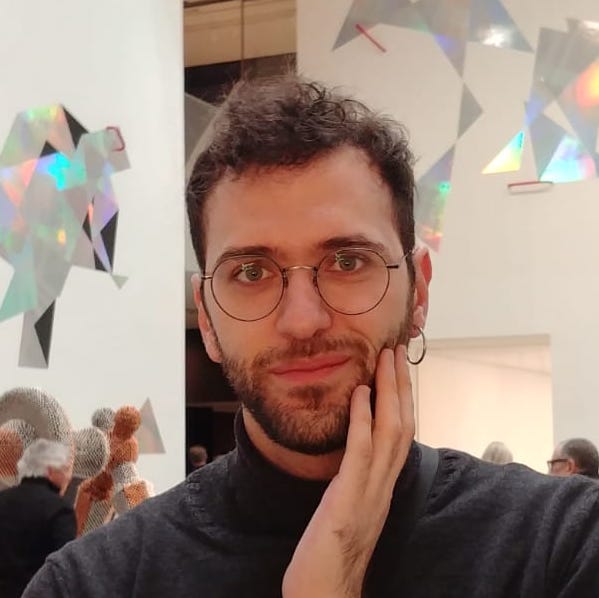}}]{Andrea Morichetta} Andrea Morichetta is a postdoctoral researcher at the Distributed Systems Group of TU Wien, specializing in machine learning, distributed systems, and edge-to-cloud computing. Dr. Morichetta holds a Ph.D. in Electrical, Electronics, and Communication Engineering from Politecnico di Torino and possesses a strong international profile. 
He has collaborated with leading institutions and industry partners such as Cisco (San Jose, CA, US), AIT (Vienna, AT), Futurewei Technologies (Seattle, WA, US), and Tsinghua University (Beijing, CN). 
With extensive teaching experience, Dr. Morichetta lectures in Distributed Systems at TU Wien, where he leads both Bachelor and Master-level courses. His research is contributing to the study of complex, large-scale distributed systems, through modular, multi-modal coordination.
\end{IEEEbiography}

\begin{IEEEbiography}[{\includegraphics[width=0.8in, keepaspectratio]{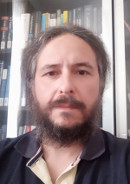}}]{Pantelis A. Frangoudis} is a researcher with the Distributed Systems Group, TU Wien, Austria. He has been with the Communication Systems Department, EURECOM, France (2017–2019), and with team DIONYSOS at IRISA/INRIA Rennes, France (2012–2017), which he originally joined under an ERCIM ``Alain Bensoussan'' post-doctoral fellowship. He has a Ph.D. (2012) in Computer Science from AUEB, Greece. His interests include mobile and wireless networking, edge and cloud computing, and Internet multimedia.
\end{IEEEbiography}

\begin{IEEEbiography}
[{\includegraphics[width=0.8in,keepaspectratio,trim=2.3in 0in 2.3in 0in,clip]{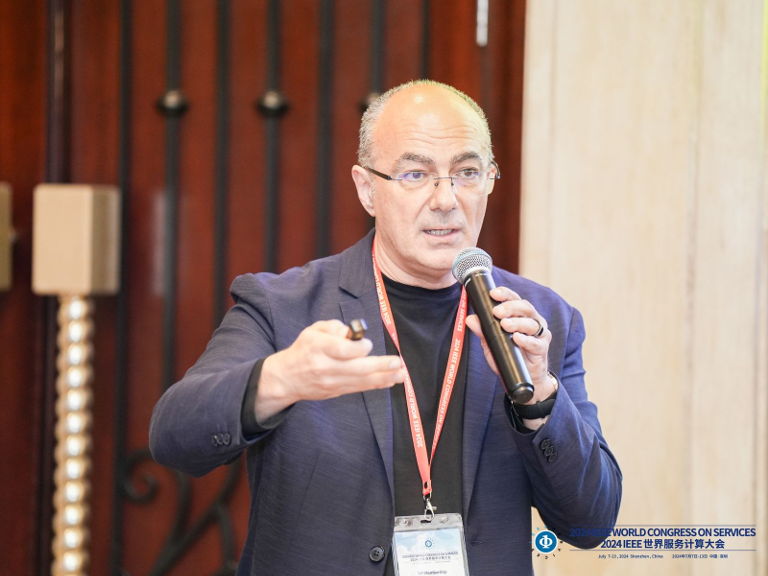}}]{Schahram Dustdar} (Fellow, IEEE) is a Full Professor of Computer Science (Informatics) with a focus on Internet Technologies heading the Distributed Systems Group at the TU Wien. 
He is Editor-in-Chief of Computing (Springer) and an Associate Editor of IEEE Transactions on Services Computing, IEEE Transactions on Cloud Computing, ACM Computing Surveys, ACM Transactions on the Web, and ACM Transactions on Internet Technology, as well as on the editorial board of IEEE Internet Computing and IEEE Computer. 
He is a recipient of multiple awards: TCI Distinguished Service Award (2021), IEEE TCSVC Outstanding Leadership Award (2018), IEEE TCSC Award for Excellence in Scalable Computing (2019), ACM Distinguished Scientist (2009), ACM Distinguished Speaker (2021), IBM Faculty Award (2012).
He is a Member of Academia Europaea und IEEE Fellow.
\end{IEEEbiography}